\documentclass[runningheads]{llncs}
\usepackage{graphicx}

\usepackage{booktabs}
\usepackage[hidelinks]{hyperref}
\usepackage{cleveref}
\usepackage[absolute,overlay]{textpos}

\def\datasetname/{\mbox{NewsTSC}}

\begin{document}

\title{Towards Target-dependent Sentiment Classification in News Articles}
%
%
\author{Felix Hamborg\inst{1,2}\orcidID{0000-0003-2444-8056} \and
Karsten Donnay\inst{2,3}\orcidID{0000-0002-9080-6539} \and
Bela Gipp\inst{2,4}\orcidID{0000-0001-6522-3019}}
\authorrunning{F. Hamborg et al.}
%
\institute{
Dept. of Computer Science, University of Konstanz, Germany \email{felix.hamborg@uni-konstanz.de} \and
Heidelberg Academy of Sciences and Humanities, Germany \and 
Dept. of Political Science, University of Zurich, Switzerland  \and
Data and Knowledge Engineering, University of Wuppertal, Germany}

\maketitle              
%
\begin{abstract}
Extensive research on target-dependent sentiment classification (TSC) has led to strong classification performances in domains where authors tend to explicitly express sentiment about specific entities or topics, such as in reviews or on social media. We investigate TSC in news articles, a much less researched domain, despite the importance of news as an essential information source in individual and societal decision making. This article introduces \datasetname/, a manually annotated dataset to explore TSC on news articles. Investigating characteristics of sentiment in news and contrasting them to popular TSC domains, we find that sentiment in the news is expressed less explicitly, is more dependent on context and readership, and requires a greater degree of interpretation. In an extensive evaluation, we find that the current state-of-the-art in TSC performs worse on news articles than on other domains (average recall $AvgRec=69.8$ on \datasetname/ compared to $AvgRev=\left[ 75.6, 82.2\right] $ on established TSC datasets). Reasons include incorrectly resolved relation of target and sentiment-bearing phrases and off-context dependence. As a major improvement over previous news TSC, we find that BERT's natural language understanding capabilities capture the less explicit sentiment used in news articles.

\end{abstract}

\begin{textblock*}{\paperwidth}(0cm,1cm) 
\centering
\noindent
   This is a preprint. The accepted manuscript can be found at: \\ \url{https://doi.org/10.1007/978-3-030-71305-8_12}
\end{textblock*}

\section{Introduction}
Target-dependent sentiment classification (TSC) is a sub-task of sentiment analysis that aims to identify the sentiment of a text, usually on sentence-level, towards a given target, such as named entities (NEs) or other semantic concepts \cite{Jiang2011Target-dependentClassification}. Aspect-based sentiment classification (ABSC) \cite{Rosenthal2017SemEval-2017Twitter}, a closely related task, defines such targets as aspects of a given topic, e.g., ``service'' and ``food'' may be aspects of the topic ``restaurant.'' Previous research on TSC and ABSC (due to their technical similarity we will refer to both as TSC) has focused mostly on domains in which authors tend to express their opinions explicitly, such as reviews, surveys, and social media \cite{Dong2014,Pontiki2015SemEval-2015Analysis,Nakov2016SemEval-2016Twitter,Rosenthal2017SemEval-2017Twitter}.

In this paper, we investigate TSC in the domain of news articles -- a much less researched domain that is of critical relevance, especially in times of ``fake news,'' echo chambers, and news owner centralization \cite{Hamborg2018e}. How persons and other entities are portrayed in articles on political topics is, e.g., very relevant for individual and societal opinion formation \cite{bernhardt2008political,Hamborg2020Newsalyze:Bias,Hamborg2020MediaLabeling}.

The main contributions of this paper are: (1) We introduce \emph{\datasetname/}, a manually annotated dataset for the exploration of TSC in political news articles. (2) We discuss similarities and differences between political news and established TSC domains. (3) We perform an extensive evaluation of state-of-the-art TSC approaches on \datasetname/. To improve classification performance, we also fine-tune a BERT language model \cite{Devlin2019BERT:Understanding} on a large news dataset, thereby establishing the current state-of-the-art in TSC on political news.

We provide the dataset including code book, code to reproduce our experiments, and the fine-tuned BERT at: \url{https://github.com/fhamborg/newstsc}

\section{Related Work}
\label{sec2}
Most TSC-related research uses three annotated datasets: \emph{Restaurant} and \emph{Laptop}, containing reviews on restaurants and laptops \cite{Pontiki2015SemEval-2015Analysis}, and \emph{Twitter}, consisting of tweets \cite{Dong2014}. Each example in these datasets consists of a \textit{target}, \textit{context} (often a single sentence or tweet), and the target's \textit{sentiment} within its context. 

The advent of word embeddings and deep learning including neural language models, such as BERT \cite{Devlin2019BERT:Understanding}, has led to a performance leap in many natural language processing (NLP) disciplines including TSC, where, e.g., macro F1 gained from $F1_m= 63.3$ \cite{Kiritchenko2015} to $F1_m= 75.8$ on the Twitter set \cite{Zeng2019LCF:Classification}. Whereas traditional TSC research focused on careful feature engineering and dictionary creation (cf. \cite{Kiritchenko2015}), researchers now focus on designing neural architectures suited to catch the relation between target and context \cite{Zhaoa2019ModelingClassification,Song2019TargetedNetwork,Zeng2019LCF:Classification}. By fine-tuning the underlying language model for the particular classification domain, performance can be improved further \cite{Rietzler2019}.

Text in news articles differs from reviews and social media in that news authors typically do not express sentiment towards a target explicitly (exceptions include opinion pieces and columns). Instead, they implicitly or indirectly express sentiment because language in news is expected to be neutral and journalists to be objective \cite{balahur2013sentiment,godbole2007large,Hamborg2018e}. For example, news texts express sentiment by describing actions performed by a target, or by including and highlighting information in favor or against a target (or omitting and downplaying such information, respectively) \cite{Schreier2012}. Adding to the difficulty of news TSC, different readers may assess an article's sentiment towards a target differently \cite{balahur2013sentiment}, depending on their own political or ideological views (we discuss real-world examples in Section~\ref{sec3.3}). Previous news TSC approaches mostly employ manually created \cite{balahur2013sentiment} or semi-automatically extended \cite{godbole2007large} sentiment dictionaries. To our knowledge, there exist one dataset for evaluation of news TSC methods \cite{Steinberger2017Large-scaleAnalysis}, which -- perhaps due to its small size ($N=1274$) -- has not been used or tested in recent TSC literature. Another dataset contains quotes extracted from news articles, since quotes more likely contain explicit sentiment ($N=1592$) \cite{balahur2013sentiment}.

To our knowledge, no suitable datasets for news TSC exist nor have news TSC approaches been proposed that exploit recent advances in NLP. 

\section{Dataset}
\label{sec3}
We describe how we create the news TSC dataset, including the collection of articles and the annotation procedure. Afterward, we discuss the characteristics of the dataset.

\subsection{Data Collection and Example Extraction}
We create a base set of articles of high diversity in topics covered and writing styles, e.g., whether emotional or factual words are used (cf. \cite{Gebhard2020ThePopularity}). Using a news extractor \cite{Hamborg2017a}, we collect news articles from the Common Crawl news crawl (CCNC, also known as CC-NEWS), consisting of over 250M articles until August 2019 \cite{Nagel2016}. To ensure diversity in writing styles, we select 14 US news outlets,\footnote{BBC, Breitbart, Chicago Tribune, CNN, LA Daily News, Fox News, HuffPost, LA Times, NBC, NY Times, Reuters, USA Today, Washington Post, and Wall Street Journal.} which are mostly major outlets that represent the political spectrum from left to right, based on selections by \cite{Budak2016FairAnalysis,groseclose2005measure,Baum2008NewDiscourse}. We cannot simply select the whole corpus, because CCNC lacks articles for some outlets and time frames. By selecting articles published between August 2017 and July 2019, we minimize such gaps while covering a time frame of two years, which is sufficiently large to include many diverse news topics. To facilitate the balanced contribution of each outlet and time-range we perform binning: we create 336 bins, one for each outlet and month, and randomly draw 10 articles reporting on politics for each bin, resulting in 3360 articles in total.\footnote{To classify whether an article reports on politics, we use a DistilBERT-based \cite{Sanh2019DistilBERTLighter} classifier with a single dense layer and softmax trained on the HuffPost \cite{Misra2018NewsDataset} and BBC datasets \cite{greene2006practical}. During the subsequent manual annotation, coders discard remaining, non-political articles.} During binning, we remove any article duplicates by text equivalence.

To create examples for annotation, we select all mentions of NEs recognized as PERSON, NROP, or ORG for each article \cite{Weischedel2013OntoNotes5.0}.\footnote{For this task, we use spaCy v2.1.} We discard NE mentions in sentences shorter than 50 characters. For each NE mention, we create an example by using the mention as the target and its surrounding sentence as its context. We remove any example duplicates. Afterward, to ensure diversity in writing styles and topics, we use the outlet-month binning described previously and randomly draw examples from each bin.

Different means may be used to address expected class imbalance, e.g., for the Twitter set, only examples that contained at least one word from a sentiment dictionary were annotated \cite{Nakov2013SemEval-2013Twitter,Nakov2016SemEval-2016Twitter}. While doing so yields high frequencies of classes that are infrequent in real-world distribution, it also causes dataset shift and selection bias \cite{Quionero-Candela2009DatasetLearning}. Thus, we instead investigate the effectiveness of different means to address class imbalance during training and evaluation (see Section~\ref{sec4}). 

\subsection{Annotation}
\label{sec3:annotation}
We set up an annotation process following best practices from the TSC literature \cite{Pontiki2015SemEval-2015Analysis,Nakov2016SemEval-2016Twitter,Rosenthal2017SemEval-2017Twitter,Steinberger2017Large-scaleAnalysis}. For each example, we asked three coders to read the context, in which we visually highlighted the target and assess the target's sentiment. Examples were shown in random order to each coder. Coders could choose from \textit{positive}, \textit{neutral}, and \textit{negative} polarity, whereby they were allowed to choose positive and negative polarity at the same time. Coders were asked to \textit{reject} an example, e.g., if it was not political or a meaningless text fragment. Before, coders read a code book that included instructions on how to code and already annotated examples. Five coders, students, aged between 24 and 32, participated in the process. 

In total, 3288 examples were annotated, from which we discard 125 (3.8\%) that were rejected by at least one coder, resulting in 3163 non-rejected examples. From these, we discard 3.3\% that lacked a majority class, i.e., examples where each coder assigned a different sentiment class, and 1.8\% that were annotated as positive and negative sentiment at the same time, to allow for better comparison with previous TSC datasets and methods (see Section \ref{sec2}). Lastly, we split the remaining 3002 examples into 2301 training and 701 test examples. Table~\ref{tab:datasetstats} shows class frequencies of the sets.

We use the full set of 3163 non-rejected examples to illustrate the degree of agreement between coders: 3.3\% lack a majority class, for 62.7\%, two coders assigned the same sentiment, and for 33.9\% all coders agreed. On average, the accuracy of individual coders is $acc_h=72.9\%$. We calculate two  intercoder reliability (ICR) measures. For completeness, Cohen's Kappa is $\kappa = 25.1$, but it is unreliable in our case due to Kappa's sensitivity to class imbalance \cite{Cicchetti1990HighParadoxes}. The mean pairwise observed agreement over all coders is $72.5$. 

\begin{table}
    \centering
    \small
    \begin{tabular}{lrrrr}
        \toprule 
        \textbf{} & \textbf{negative} & \textbf{neutral} & \textbf{positive} & \textbf{total} \\ 
        \midrule
        \textbf{training} & 530 & 1600 & 171 & 2301 \\
        \textbf{test} & 167 & 487 & 47 & 701 \\
        \midrule
        \textbf{total} & 697 & 2087 & 218 & 3002 \\
        \bottomrule
    \end{tabular}
    \caption{\label{tab:datasetstats} Class frequencies of \datasetname/ sets.}
\end{table}

\subsection{Characteristics of Sentiment in News Articles}
\label{sec3.3}
In a manual, qualitative analysis of \datasetname/, we find two key differences of news compared to established domains: first, we confirm that news contains mostly implicit and indirect sentiment (see Section~\ref{sec2}). Second, determining the sentiment in news articles typically requires a greater degree of interpretation (cf. \cite{Steinberger2017Large-scaleAnalysis}). The second difference is caused by multiple factors, particularly the implicitness of sentiment (mentioned as the first difference) and that sentiment in news articles is more often dependent on non-local, i.e., off-sentence, context. In the following, we discuss annotated examples (part of the dataset and discarded examples) to understand the characteristics of target-dependent sentiment in news texts.

We find that in news articles, a key means to express targeted sentiment is to describe actions performed by the target. This is in contrast, e.g., to product reviews where more often a target's feature, e.g., ``high resolution'', or the mention of the target itself, e.g., ``the camera is awesome,'' express sentiment. For example, in ``The \underline{Trump administration} has worked tirelessly to impede a transition to a green economy with actions ranging from opening the long-protected Arctic National Wildlife Refuge to drilling, [...].'' the target (underlined) was assigned negative sentiment due to its actions. 

We find sentiment in $\approx 3\%$ of the examples to be strongly reader-dependent (cf. \cite{balahur2013sentiment}).\footnote{We drew a random sample of 300 examples and concluded in a two-person discussion that the sentiment in 8 examples could be perceived differently.} In the previous example, the perceived sentiment may, in part, depend on the reader's own ideological or political stance, e.g., readers focusing on economic growth could perceive the described action positively whereas those concerned with environmental issues would perceive it negatively.

In some examples, targeted sentiment expressions can be interpreted differently due to ambiguity. As a consequence, we mostly find such examples in the discarded examples and thus they are not contained in \datasetname/. While this can be true for any domain (cf. ``polarity ambiguity'' in \cite{Pontiki2015SemEval-2015Analysis}), we think it is especially characteristic for news articles, which are lengthier than tweets and reviews, giving authors more ways to refer to non-local statements and to embed their arguments in larger argumentative structures. For instance, in ``And it is true that even when using similar tactics, \underline{President Trump} and President Obama have expressed very different attitudes toward immigration and espoused different goals.'' the target was assigned neutral sentiment. However, when considering this sentence in the context of its article \cite{Taub2018HowBorder}, the target's sentiment may be shifted (slightly) negatively. 

From a practical perspective, considering more context than only the current sentence seems to be an effective means to determine otherwise ambiguous sentiment expressions. By considering a broader context, e.g., the current sentence and previous sentences, annotators can get a more comprehensive understanding of the author's intention and the sentiment the author may have wanted to communicate. The greater degree of interpretation required to determine non-explicit sentiment expressions may naturally lead to a higher degree of subjectivity. Due to our majority-based consolidation method (see Section~\ref{sec3:annotation}), examples with non-explicit or apparently ambiguous sentiment expressions are not contained in \datasetname/. 

\section{Experiments and Discussion}
\label{sec4}
We evaluate three TSC methods that define the state-of-the-art on the established TSC datasets Laptop, Restaurant, and Twitter: AEN-BERT \cite{Song2019TargetedNetwork}, BERT-SPC \cite{Devlin2019BERT:Understanding}, and LCF-BERT \cite{Zeng2019LCF:Classification}. Additionally, we test the methods using a domain-adapted language model, which we created by fine-tuning BERT (base, uncased) for 3 epochs on 10M English sentences sampled from CCNC (cf. \cite{Rietzler2019}). For all methods, we test hyperparameter ranges suggested by their respective authors.\footnote{Epochs $\in \{ 3,4 \}$; batch size $\in \{16, 32\}$; learning rate $\in \{ 2e-5, 3e-5, 5e-5 \}$; label smoothing regularization (LSR) \cite{Szegedy2016RethinkingVision}: $\epsilon \in \{ 0, 0.2$ \}; dropout rate: $0.1$; $\mathcal{L}_2$ regularization: $\lambda = 10^{-5}$. We use Adam optimization \cite{Kingma2014Adam:Optimization}, Xaviar uniform initialization \cite{Glorot2010UnderstandingNetworks}, and cross-entropy loss \cite{Goodfellow2016DeepLearning}. Where multiple values for a hyperparameter are given, we test all their combinations in an exhaustive search.} Additionally, we investigate the effects of two common measures to address class imbalance: weighted cross-entropy loss (using inverse class frequencies as weights) and oversampling of the training set. Of the training set, we use 2001 examples for training and 300 for validation.

We use average recall ($AvgRec$) as our primary measure, which was also chosen as the primary measure in the TSC task of the latest SemEval series, due to its robustness against class imbalance \cite{Rosenthal2017SemEval-2017Twitter}. We also measure accuracy ($acc$), macro F1 ($F1_m$), and average F1 on positive and negative classes ($F1_{pn}$) to allow comparison to previous works \cite{Nakov2016SemEval-2016Twitter}. 

Table \ref{table:results} shows that LCF-BERT performs best ($AvgRec=67.3$ using BERT and $69.8$ using our news-adapted language model).\footnote{Each row in Table~\ref{table:results} shows the results of the hyperparameters that performed best on the validation set.} Class-weighted cross-entropy loss helps best to address class imbalance ($AvgRec=69.8$ compared to $67.2$ using oversampling and $64.6$ without any measure).

\begin{table}
    \small
    \centering
    \begin{tabular}{lcrrrr}
    \toprule
    \textbf{LM} & \textbf{Method} & \textbf{AvgRec} & \textbf{acc} & $\mathbf{F1_m}$ & $\mathbf{F1_{pn}}$ \\ 
    \midrule
         & AEN-BERT & 59.7 & 62.9 & 55.0 & 47.3 \\
    base & BERT-SPC & 62.1 & 62.1 & 53.3 & 44.9 \\
         & LCF-BERT & \textbf{67.3} & 61.3 & 54.4 & 46.5 \\
    \midrule
         & AEN-BERT & 59.8 & 62.9 & 54.5 & 46.2 \\
    news & BERT-SPC & 66.7 & 63.5 & 55.0 & 45.8 \\
         & LCF-BERT & \textbf{69.8} & 66.0 & 58.8 & 51.4 \\
    \bottomrule
    \end{tabular}
    \caption{\label{table:results}Experiment results. \textit{LM} refers to the language model used, where \textit{base} is BERT (base, uncased) and \textit{news} is our fine-tuned BERT model.}
\end{table}

Performance in news articles is significantly lower than in established domains, where the top model (LCF-BERT) yields in our experiments $AvgRev=78.0$ (Laptop), $82.2$ (Restaurant), and $75.6$ (Twitter). For Laptop and Restaurant, we used domain-adapted language models \cite{Rietzler2019}. News TSC accuracy $acc=66.0$ is lower than single-human-level $acc_h=72.9$ (see Section~\ref{sec3.3}).

We carry out a manual error analysis (up to 30 randomly sampled examples for each true class). We find \textit{target misassociation} as the most common error cause: in 40\%, sentences express the predicted sentiment towards a different target. In 30\%, we cannot find any apparent cause. The remaining cases contain various potential causes, including usage of euphemisms or sayings (12\% of examples with negative sentiment). Infrequently, we find that sentiment is expressed by rare words or figurative speech, or is reader-dependent (the latter in 2\%, approximately matching the 3\% of reader-dependent examples reported in Section~\ref{sec3.3}).

Previous news TSC approaches, mostly dictionary-based, could not reliably classify implicit or indirect sentiment expressions (see Section \ref{sec2}). In contrast, our experiments indicate that BERT's language understanding suffices to interpret implicitly expressed sentiment correctly (cf. \cite{Devlin2019BERT:Understanding,balahur2013sentiment,godbole2007large}). \datasetname/ does not contain instances in which the broader context defines sentiment, since human coders could not classify them correctly in the first place. Our experiments therefore cannot elucidate this particular characteristic discussed in Section~\ref{sec3.3}.

\section{Conclusion and Future Work}
We explore how target-dependent sentiment classification (TSC) can be applied to political news articles. Our main contributions are as follows: first, we introduce \datasetname/, a dataset to explore target-dependent sentiment classification (TSC) in political news articles, consisting of over 3000 manually annotated sentences. 

Second, in a qualitative analysis, we find notable differences concerning how authors express sentiment towards targets as compared to other well-researched domains of TSC, such as product review or posts on social media. In these domains, authors tend to explicitly express their opinions. In contrast, in news articles, we find dominant use of implicit or indirect sentiment expressions, e.g., by describing actions, which were performed by a given target, and their consequences. Thus, sentiment expressions may be more ambiguous, and determining their polarity requires a greater degree of interpretation. 

Third, in a quantitative evaluation, we find that state-of-the-art TSC methods perform lower on the news domain (average recall $AvgRec=69.8$ using our news-adapted BERT model, $AvgRec=67.3$ without) than on popular TSC domains ($AvgRec=\left[ 75.6, 82.2\right]$). 

We identify multiple future research directions for news TSC. While \datasetname/ contains clear sentiment expressions, it lacks other sentiment types that occur in real-world news coverage. For example, sentences that express sentiment more implicitly or ambiguously. To create a labeled TSC dataset that better reflects real-world news coverage, we suggest to adjust annotation instructions to raise annotators' awareness of these sentiment types and clearly define how they should be labeled. Technically, apparently ambiguous sentiment expressions might be easier to label when considering a broader context, e.g., not only the current sentence but also previous sentences. Considering more context might also help to improve a classifier's performance. 

We envision to integrate TSC methods into a system that identifies slanted news coverage \cite{Hamborg2019AutomatedArticles,Spinde2020EnablingBias}. For example, given a set of articles reporting on the same topic, a system could identify articles that similarly frame the actors involved in the event. To do so, the system would analyze frequently mentioned persons' polarities in each article. Then, it would group articles that similarly portray these persons.

\subsubsection*{Acknowledgements}
The work described in this paper is partially funded by the WIN program of the Heidelberg Academy of Sciences and Humanities, financed by the Ministry of Science, Research and the Arts of the State of Baden-Wurttemberg, Germany. The authors thank the students who participated in the manual annotation as well as the anonymous reviewers for their valuable comments.

\bibliographystyle{splncs04}
\bibliography{short}

\end{document}